\documentclass[letterpaper, 10 pt, conference]{ieeeconf}  

\usepackage{hyperref}
\usepackage{amsmath}
\usepackage{amsfonts}
\usepackage{wrapfig}
\usepackage{xcolor}
\usepackage{graphics} 
\usepackage{epsfig} 
\usepackage{algorithm2e}
\RestyleAlgo{ruled}

\hypersetup{
  colorlinks   = true, 
  urlcolor     = red, 
  linkcolor    = blue, 
  citecolor   = blue 
}

\IEEEoverridecommandlockouts                              

\title{\LARGE \bf
Closing the Visual Sim-to-Real Gap with Object-Composable NeRFs
}

\author{Nikhil Mishra$^{*, 1, 2}$,  Maximilian Sieb$^{3}$, Pieter Abbeel$^{1, 2}$, Xi Chen$^{1}$
\thanks{$^*$Correspondence to: nmishra@berkeley.edu.}%
\thanks{
$^{1}$Covariant.ai, 
$^{2}$UC Berkeley, 
$^{3}$Work done at Covariant.ai}%
}

\begin{document}

\maketitle
\thispagestyle{empty}
\pagestyle{empty}

\begin{abstract}
Deep learning methods for perception are the cornerstone of many robotic systems. 
Despite their potential for impressive performance, obtaining real-world training data is expensive, and can be impractically difficult for some tasks.
Sim-to-real transfer with domain randomization offers a potential workaround, but often requires extensive manual tuning and results in models that are brittle to distribution shift between sim and real.
In this work, we introduce Composable Object Volume NeRF (COV-NeRF), an object-composable NeRF model that is the centerpiece of a real-to-sim pipeline for synthesizing training data targeted to scenes and objects from the real world.
COV-NeRF extracts objects from real images and composes them into new scenes, generating photorealistic renderings and many types of 2D and 3D supervision, including depth maps, segmentation masks, and meshes.
We show that COV-NeRF matches the rendering quality of modern NeRF methods, and can be used to rapidly close the sim-to-real gap across a variety of perceptual modalities.
\end{abstract}

\section{INTRODUCTION}

Nearly all applications in robotics require perception of the physical world, and deep learning is the method of choice for nearly all tasks in computer vision.
As neural network architectures and training recipes have matured, limits on training data have become the bottleneck for these methods.
Even if a large dataset is available, deployed systems may encounter scenarios that their original training did not prepare them for.
Moreover, while some tasks can be supervised via human annotation (for example, object detection or semantic segmentation), other tasks, like depth estimation or shape completion, are best supervised in simulation.

However, when models trained in simulation are transferred  to the real-world, their performance often degrades because the input distribution has shifted (a well-documented phenomenon known as the sim-to-real gap).
Domain randomization \cite{domain_randomization} is typically used to improve robustness to out-of-distribution inputs: if a model is trained to generalize to different parameters of the simulator, such as viewpoint, lighting, or material properties, then it may also generalize to the real world as simply another randomization.
However, this requires the training distribution to be both diverse enough to enable generalization, as well as realistic enough that the real world is a plausible sample from this distribution.
This can be difficult to achieve in practice, often requiring manual asset creation (meshes, textures, materials), scene construction, and parameter tuning.
Moreover, if a large sim-to-real gap is observed, it can be difficult to know exactly how to improve the simulator, which may require specialized expertise (e.g. graphics for visual tasks) or domain/task-specific tuning.

\begin{figure}[t!]
  \centering
  \includegraphics[width=0.48\textwidth]{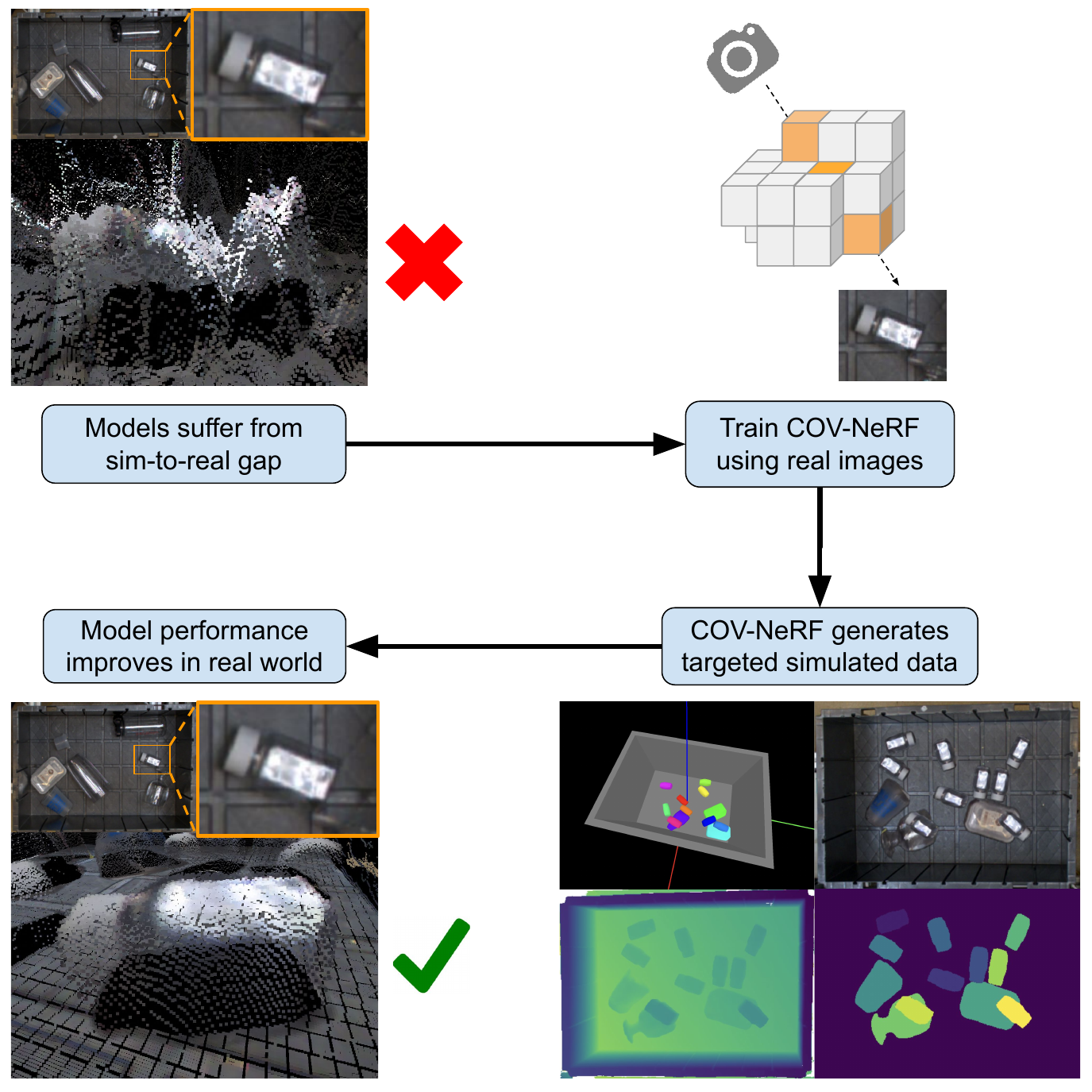}
  \caption{
  Our proposed object-centric neural renderer, COV-NeRF, can be used to generate targeted supervision for other models that are brittle to sim-to-real distribution shift.
  After learning explicit neural representations of real objects, COV-NeRF can compose those representations into photorealistic synthetic scenes and generate many modalities of downstream supervision, including depth maps, segmentation masks, and instance meshes.
  }
  \label{figure_overview}
  \vspace{-5mm}
\end{figure}

In this work, we explore neural rendering methods as a mechanism to close the sim-to-real gap for perception tasks.
The seminal NeRF \cite{nerf} uses real images to build a neural representation of a scene, which can then be rendered from new viewpoints through a ray-marching process based on physical models of light transport.
However, these methods typically require extensive test-time optimization (TTO), taking many GPU-hours to reconstruct a single scene.
In this work, we introduce Composable Object Volume NeRF, or COV-NeRF, a novel neural renderer that is uniquely suited to remedy sim-to-real mismatch.
Unlike prior NeRF methods, COV-NeRF explicitly represents objects while also generalizing across scenes without TTO.
COV-NeRF performs real-to-sim learning of structured neural representations that can subsequently be used for novel scene synthesis, rendering photorealistic images along with the corresponding supervision for many perception tasks.
Existing methods for domain adaptation simply re-skin simulated scenes \cite{cycada} \cite{retinagan}; we synthesize new scenes \textit{de novo}, can produce both RGB images and many types of 2D and 3D supervision, and automatically enjoy geometric and semantic consistency across viewpoints and between the rendered images and labels.

Our key contributions are as follows:
\begin{enumerate}
\item 
We introduce COV-NeRF, a novel NeRF architecture that both explicitly represents objects and generalizes across scenes.
Although these properties have been explored individually in prior work; COV-NeRF is the first to exhibit both simultaneously.
\item
We demonstrate COV-NeRF's ability to generate targeted synthetic data based on real-world scenes and objects.
Without artificial consistency constraints, we show that COV-NeRF can generate supervision for a variety of perceptual tasks, including depth estimation, object detection, instance segmentation, and shape completion, and that training on COV-NeRF's generations improve real-world performance.

\item 
Finally, we apply the entire pipeline to a real world bin-picking application.
We identify challenging scenarios where state-of-the-art perception models and simulated datasets face a large sim-to-real gap, and show that COV-NeRF's real-to-sim capabilities can rapidly close the gap to achieve application-level improvement.

\end{enumerate}

\section {Related Work}
\label{sec_related_work}

\textbf{Neural Rendering:}
NeRF \cite{nerf} was a breakthrough in physically-plausible differentiable rendering, yielding better visual quality and easier optimization than existing methods for inverse rendering.
However, it required 50-100 source images and several GPU-days of TTO, since it had to optimize new neural networks for each scene.
More recent methods, such as PixelNeRF \cite{pixelnerf}, MVS-NeRF \cite{mvsnerf}, and NerFormer \cite{nerformer} use more specialized architectures that can be trained simultaneously on many scenes and generalize to new scenes without TTO.
Since they learn priors that transfer across scenes, they only need 3-5 source views during inference.

Object-NeRF \cite{objectnerf} and Object Scattering Functions (OSF) \cite{osf} stayed within the single-scene paradigm, but explored object-centric decompositions that allowed scene editing, where objects can be explicitly added, re-positioned, or scaled.
OSF's edited scenes are the most realistic, but it requires the lighting to be specified parametrically (which can be difficult outside of controlled settings), and an order of magnitude more compute during rendering.
Subsequent work \cite{panoptic_nf} \cite{panoptic_lifting} adds bells-and-whistles to Object-NeRF; COV-NeRF extends these methods to not require TTO.

In robotics, single-scene NeRFs have been explored for online geometry estimation, but were bottlenecked by the computational cost of TTO \cite{dexnerf} \cite{evonerf}.
NeRF-Supervised Deep Stereo \cite{nerf_stereo} used single-scene NeRFs to generate labels for stereo depth estimation.
This pipeline generated high-quality supervision, but required extensive TTO and many source views per scene.
GraspNeRF \cite{grasp_nerf} trained NeRF rendering in simulation as an auxiliary loss for grasp generation, but did not explore or evaluate its rendering quality, let alone its ability to mitigate sim-to-real mismatch.
GraspNeRF's architecture is similar to MVS-NeRF; as such, we expect similar rendering performance, and it does not share COV-NeRF's object-composable properties.

\textbf{Sim-to-Real Transfer:}
Domain randomization has been explored for sim-to-real transfer of both visual and physical tasks.
Early visual sim-to-real work  \cite{domain_randomization} focused on diversity rather than realism: these methods achieved promising results on simple tasks and environments, but could not scale to more challenging applications without manual tuning.
Simple heuristics like cut-and-paste were explored for segmentation, but produce visually and geometrically implausible scenes, and cannot generate 3D supervision  \cite{cutpaste} \cite{data_dreaming}.

Domain adaptation methods employed generative models to automatically improve the realism of simulated images.
CyCADA \cite{cycada} trained GANs to translate simulated images into real ones, but required a cycle consistency objective so that the original semantic segmentations could still serve as supervision for the translated images.
Subsequent work applied this idea using object detection \cite{retinagan} and Q-values \cite{rlcyclegan}.
However, finding such a bijective mapping is an ill-posed task, and its difficulty scales poorly as more modalities are considered (e.g. if we require consistency for depth maps as well as for semantic segmentation).
We take a fundamentally different approach: COV-NeRF creates datasets \textit{de novo} that are more realistic and more meaningfully diverse than the adaptions produced by these methods, and achieves multi-view consistency for many modalities for free.
We evaluate domain adaptation methods in Section \ref{sec_sim2real_experiments}.

Inspired by recent advances in vision-language alignment, CACTI \cite{cacti}, ROSIE \cite{rosie}, and GenAug \cite{genaug} used text-conditioned diffusion models \cite{ddpm} to automatically apply augmentations, such as changing backgrounds or adding distractor objects.
These methods improved the robustness of end-to-end policies, but ultimately suffer from the same limitations as the GAN-based approaches.

\begin{figure*}[t!]
  \centering
  \includegraphics[width=1.0\textwidth]{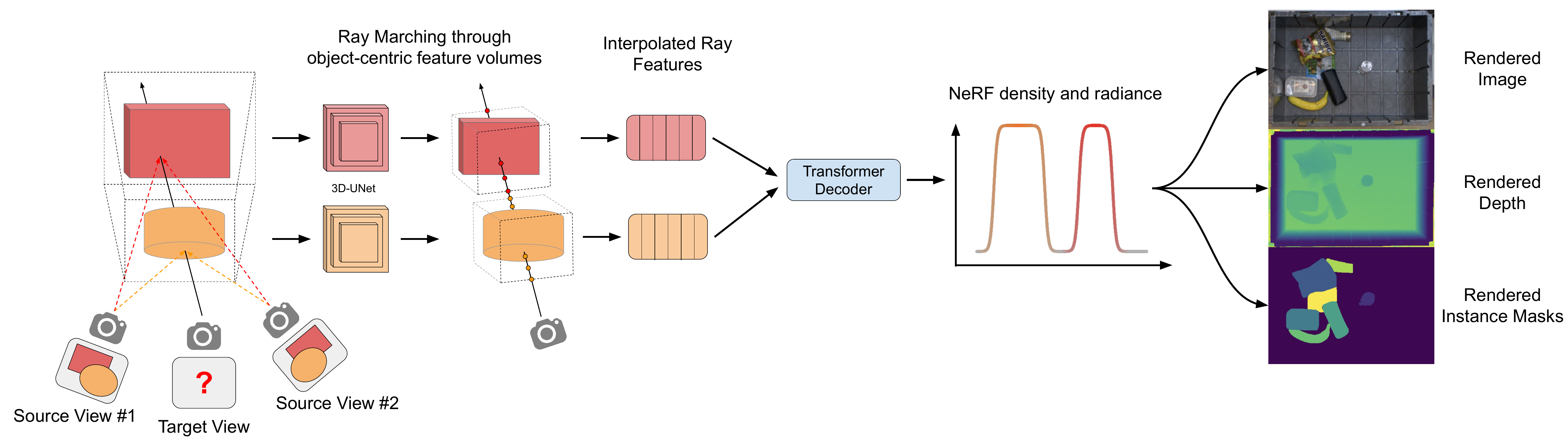}
  \vspace{-6mm}
  \caption{
  An overview of COV-NeRF's object-centric rendering process.
  Visual features from the source views are projected into a feature volume for each object in the scene.
  For each pixel to be rendered, features are interpolated along the corresponding ray from each volume that the ray intersects with.
  A Transformer decodes the interpolated features into the NeRF density and radiance, which are composited into RGB colors, depths and segmentation masks.
  }
  \label{figure_rendering}
  \vspace{-5mm}
\end{figure*}

\section{Generalizable Neural Rendering with Composable Object Volumes}

\subsection{Preliminaries}

NeRF methods use differentiable volumetric rendering to train models for novel view synthesis.
Given a collection of source images $\{ \mathcal{I}^{(b)} \}_{b=1}^{B} $ of a scene and their corresponding camera intrinsic/extrinsic matrices, their goal is to render an image $\mathcal{I}^*$ of the scene as seen from a new viewpoint.

Recall that a ray $\mathbf{r}$ is parametrized as $\mathbf{r}(t) = o + t \cdot d, t \geq 0$, where $o, d \in \mathbb{R}^3$ are the ray's origin and direction.
Each pixel $(u,v) \in \mathcal{I}^*$ corresponds to a particular ray $\mathbf{r}_{u,v}(t)$; the intrinsics and extrinsics determine $o$ and $d$.
To render a ray, points $\mathbf{r}(t_i), t_1 < \dots < t_N$ are sampled along it, and the density $\sigma_i$ and radiance $c_i$ are determined for each, then accumulated to compute the ray's rendered color $\hat C(\mathbf{r})$:
\begin{align}
\label{eq_volumetric_rendering}
\hat C(\mathbf{r}) 
&= \sum_{i=1}^{N} T_i  \alpha_i  c_i
\end{align}
where $\delta_i = t_{i+1} - t_i$, $\alpha_i = 1 - e^{ -\delta_i \sigma_i } $, $T_i = e^{ -\sum_{j<i} \delta_j \sigma_j }$.
The rendered color of pixel $(u, v)$ is simply $\hat C(\mathbf{r}_{u,v}(t))$.

Early NeRFs used fully-connected neural networks to predict $\sigma_i$ and $c_i$ from the coordinate $\mathbf{r}_{u,v}(t_i))$ and the ray direction $d$.
However, since they lacked direct access to the source images during inference, they had to distill visual details from the source images into the network weights via optimization, which prevented them from representing multiple scenes.
Moreover, since the networks were trained from scratch for each scene, a large $B \approx 100$ was required.

In the scene-generalizable extensions discussed in Section \ref{sec_related_work}, the dependence on $\{ \mathcal{I}^{(b)} \} $ was made explicit by making these images inputs to the network during rendering.
More powerful architectures, such as 3D-CNNs and Transformers, were introduced to better leverage the additional information.
Instead of memorizing visual details, the networks learned to robustly aggregate information from the source images, and could work with substantially fewer source views ($B \approx 3$).

\subsection{Composable Object Volumes}

For synthetic data generation, we seek a NeRF method that both generalizes across scenes and explicitly represents objects.
These properties ensure that we can generate supervision for different perceptual tasks, that the generated supervision is consistent with the rendered images, and that we can scalably generate diverse datasets with many objects in a variety of configurations.

COV-NeRF represents a scene as a collection of objects $ \{o_k \}_{k=1}^{K}$ and a background.
Each object is defined by a frustum, which is discretized into a $D \times H \times W$ voxel grid.
A learned feature vector is associated with each voxel, resulting in a feature volume $V^{(k)} \in \mathbb{R}^{C \times D \times H \times W}$.
The object $o^{(k)}$ is fully specified by its pose $p^{(k)} \in \mathcal{SO}^3$ and volume $V^{(k)}$.

Each $V^{(k)}$ is computed by projecting the RGB values from the source views, resulting in an initial volume of shape $B \small{\times} 3 \small{\times}D \small{\times}H \small{\times} W$.
We aggregate features across source views via attention over the $B$ dimension, reducing the volume to $C \small{\times} D \small{\times} H \small{\times} W$, and then refine it with a 3D-UNet to produce $V^{(k)}$.
MVS-NeRF \cite{mvsnerf} uses a similar plane-sweep scheme, but constructs one volume for the entire scene, while we construct one for each object.

To render a ray, we gather contributions from each object and the background.
For the background, we sample points $r(t_i^{(0)}), t^{(0)}_1, \dots, t^{(0)}_N$ throughout the entire scene, and predict $\sigma_i^{(0)}, c_i^{(0)}$ using a simplified NerFormer \cite{nerformer}, which interpolates features from each source view and processes them with a Transformer.
For each object $o^{(k)}$, if the ray intersects with its frustum, we also sample points $r(t_i^{(k)}), t^{(k)}_1, \dots, t^{(k)}_N$.
We transform $r(t_i^{(k)})$ and $d$ into the object frame $p^{(k)}$, trilinearly interpolate feature vectors from $V^{(k)}$, and use a Transformer to decode them into $\sigma_{i}^{(k)}$ and $c_{i}^{(k)}$ following \cite{nerformer} \cite{neuray}.

To accumulate these values into $\hat C(\mathbf{r})$, we sort all the $\{ t_i^{(k)} \}_{i=1,k=0}^{i=N, k=K}$ and apply Equation \ref{eq_volumetric_rendering}.
This natively handles occlusions: even if $\sigma_{i}^{(k)}$ is large, $T_{i}^{(k)}$ may be small if the ray passes through other objects before hitting object $o_k$.

Following prior work, the product $\alpha_i T_i$ can be interpreted as the probability that the ray terminates at distance $t_i$.
Then the expected distance $\hat \tau(\mathbf{r})$ that the ray travels is:
\begin{align}
\label{eq_depth_rendering}
\hat \tau(\mathbf{r}) 
&= \sum_{i=1}^{N} T_i  \alpha_i t_i
\end{align}
$\hat \tau(\mathbf{r})$ is the point-to-point distance from the camera center, but can be trivially converted into a depth value in the camera frame.
COV-NeRF renders depth maps using this scheme, considering all objects and the background.

This probabilistic interpretation can also be leveraged to render instance masks.
In addition to standard (modal) instance masks, which indicate the pixels where an object is visible, COV-NeRF can also render amodal masks, which include both visible and occluded portions of each object.

Let $M^{(k)}_{u, v} \in [0, 1]$ be the probability that pixel $(u, v)$ belongs to object $o_k$'s modal mask, and let $\bar M^{(k)}_{u, v} $ be the corresponding probability for its amodal mask.
$M^{(k)}_{u, v}$ is simply the probability that $\mathbf{r}_{u, v}$ terminates inside $o_{k}$, and $\bar M^{(k)}_{u, v}$ is the probability that $\mathbf{r}_{u, v}$ would terminate inside $o^{(k)}$ in the absence of other objects:
\begin{align}
\label{eq_render_segm}
M^{(k)}_{u, v}
&= \sum_{i=1}^{N} T_i^{(k)} \alpha_i^{(k)} \\
\bar M^{(k)}_{u, v}
&= \sum_{i=1}^{N} \bar T_i^{(k)} \alpha_i^{(k)},
\text{  where  }
\bar T_{i}^{(k)} = \exp \big( -\sum_{j<i} \delta_j^{(k)} \sigma_j^{(k)} \big)
\end{align}

For more details about COV-NeRF's architecture and rendering, see \href{https://github.com/nikhilmishra000/cov-nerf}{our implementation}.

\subsection{Training COV-NeRF}

Like other scene-generalizable NeRF methods, COV-NeRF learns visual and geometric priors that enable inference when only a few source images are available.
To best learn these priors, we train COV-NeRF on a mix of simulated and real data: simulation helps bootstrap its understanding of 3D geometry, and real data exposes it to realistic textures, materials, and lighting.
This gives us the best of both worlds: COV-NeRF can leverage dense geometric supervision when available, but is also robust to sim-to-real mismatch, since it can be trained on any real data that it does not initially generalize to.
We jointly train the following losses:
\begin{itemize}
\item View synthesis: the rendered color (Equation \ref{eq_volumetric_rendering}) is trained to match the true color using an L2-loss.
\item Depth estimation: the rendered depth (Equation \ref{eq_depth_rendering}) is trained to match the true depth using an L1 loss.
\item Instance segmentation: the rendered masks (Equation \ref{eq_render_segm}) are trained to match the ground-truth masks using a cross-entropy loss.
\item Voxel occupancy: we use $V^{(k)}$ to predict the occupancy of each voxel in that object's feature volume, following \cite{meshrcnn} \cite{occupancy}.
This facilitates novel scene synthesis (see Section \ref{sec_scene_generation}) since it lets us automatically pose objects into physically plausible configurations.
\end{itemize}

In practice, we found that an efficient scheme is to pre-train all of the above losses in simulation, and add real data when it becomes available.
We pre-train for 100 epochs on COB-3D-v2 \cite{occupancy}, a small but high-quality simulated dataset that contains all of the necessary supervision modalities, which takes about 1 day using 8 NVIDIA RTX A5000s.
We used a batch size of 32 scenes and the Adam optimizer with default parameters ($\alpha = 10^{-3}, \beta_1 = 0.9, \beta_2 = 0.999$).

\subsection{Scene Generation with COV-NeRF}
\label{sec_scene_generation}

Single-scene NeRF methods (see Section \ref{sec_related_work}) anecdotally explore scene-editing as a benefit of object-centric rendering.
We extend this capability to perform novel scene synthesis, procedurally generating new scenes at scale, for use in a downstream application.
When sim-to-real models struggle in the real world, COV-NeRF can synthesize new training data targeting specific real-world scenes and objects.

Given a pre-trained COV-NeRF and collection of real scenes, we first perform real-to-sim finetuning using the captured images and any additional supervision that may be available (such as instance mask annotations).
Although this step is not strictly required, the ability to use weak 2D supervision as learning signal for the underlying 3D semantics and geometry (for which direct real-world supervision usually cannot be obtained) is a strength unique to COV-NeRF, and is only possible because of its object- and ray-centric structure.
Next, we extract feature volumes and meshes for objects from the real scenes.
The meshes are extracted from the voxel occupancy predictions via Marching Cubes.
To compose a new scene, we sample objects, pose them in geometrically realistic configurations using their meshes and a physics simulator (we use Mujoco \cite{mujoco}), and then render the scene using COV-NeRF.
The process is described in Algorithm \ref{alg_generation}.

\begin{algorithm}[t!]
\caption{Scene Generation with COV-NeRF}
\label{alg_generation}
\SetKwInOut{Input}{Input}
\Input{Desired number of scenes $N$\\
Feature volumes $\mathcal{V} = \{ V^{(1)} \dots, V^{(K)} \}$ \\
Background source images $\{\mathcal{I}^{(1)}, \dots, \mathcal{I}^{(B)}\}$
}

\For{i = 1, \dots, N}{
Sample camera viewpoints \\
Sample a number of objects $K^{(i)}$  \\
Initialize physics simulator (e.g. Mujoco) \\
\For{$k = 1, \dots, K^{(i)}$}{
    Sample object $V^{(k)} \sim \mathcal{V}$ and corresponding mesh \\
    Sample initial pose $p_0^{(k)}$ \\
    Add mesh to simulator at pose $p_0^{(k)}$ \\
}
Advance physics for $T$ timesteps (e.g. until all objects settle) \\ 
Render images and supervision with $V^{(k)}, p_T^{(k)}, k = 1, \dots, K^{(i)}$ and $\{\mathcal{I}^{(1)}, \dots, \mathcal{I}^{(B)}\}$
}
\end{algorithm}

\section{Experiments}

We conducted experiments in both simulation and the real world to answer the following questions:
\begin{enumerate}
\item How does COV-NeRF compare to other NeRF methods, in terms of the visual quality of its renderings?
\item How effective is synthetic data generated by COV-NeRF for training perception models relevant to robotic applications?
\item How effective is COV-NeRF at reducing the sim-to-real gap in challenging scenarios?
\end{enumerate}

\subsection{View Synthesis}

\begin{figure*}[t!]
  \centering
  \includegraphics[width=1.0\textwidth]{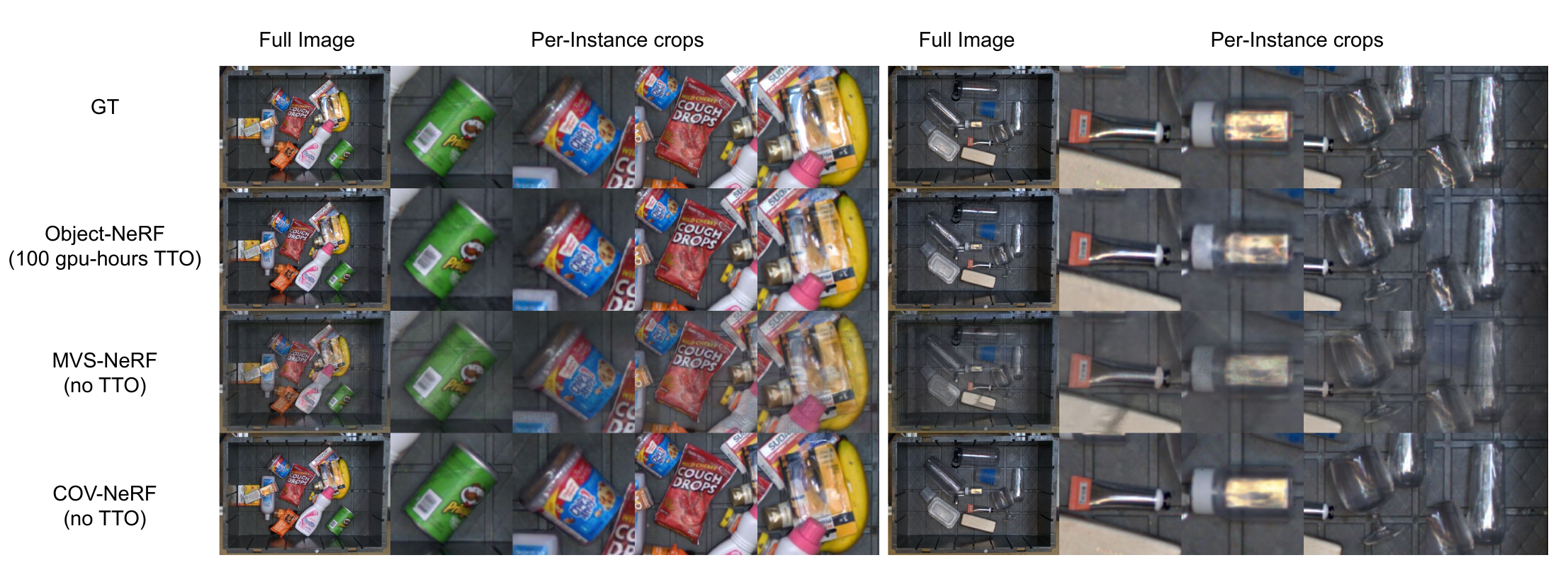}
  \vspace{-7mm}
  \caption{
  Qualitative view synthesis results on two real scenes.
  For each method, an image is rendered from a novel viewpoint using 4 source views (not pictured).
  The ground-truth image from the novel viewpoint is show in the top row.
  COV-NeRF matches the performance of object-centric methods that require expensive, per-scene TTO, and outperforms other scene-generalizable methods.
  }
  \vspace{-2mm}
  \label{figure_view_synthesis_real}
\end{figure*}

To evaluate COV-NeRF's rendering capabilities, we compare it to the following methods:
\begin{itemize}
\item 
MVS-NeRF \cite{mvsnerf} is a state-of-the-art scene-generalizable NeRF method.
It shares architectural similarities to COV-NeRF, but is not object-centric.

\item 
Object-NeRF \cite{objectnerf} is an object-centric renderer like COV-NeRF, but must be retrained for each scene. 
It requires many sources views and mask supervision in each.

\end{itemize}

\begin{table}[b!]
\caption{Scene-Generalizable View Synthesis on COB-3D-v2}
\vspace{-7mm}
\label{table_view_synthesis_cob}
\begin{center}
\resizebox{0.49\textwidth}{!}{  
    \begin{tabular}{|c|c c|c c c|}
    \hline
    Method & TTO? & Object-Centric? & PSNR $(\uparrow)$ & SSIM $(\uparrow)$ & LPIPS $(\downarrow)$ \\
    \hline
    MVS-NeRF & N & N & 25.03 & 0.867 & 0.102\\
    COV-NeRF & N & Y & \textbf{28.32} & \textbf{0.915} & \textbf{0.082} \\
    \hline
    \end{tabular}
}
\end{center}
\end{table}

In Table \ref{table_view_synthesis_cob}, we evaluate MVS-NeRF and COV-NeRF on the COB-3D-v2 validation set ($\sim$600 scenes). 
Object-NeRF is excluded due to computational constraints.
In Table \ref{table_view_synthesis_real}, we evaluate all methods on scenes from the real-world environment from Section \ref{sec_sim2real_experiments}.
In Figure \ref{figure_view_synthesis_real}, we show qualitative examples from these scenes.
COV-NeRF strictly outperforms MVS-NeRF, and, without TTO, matches or exceeds the performance of single-scene methods like Object-NeRF.

\begin{table}[h!]
\caption{View Synthesis, real}
\vspace{-7mm}
\label{table_view_synthesis_real}
\begin{center}
    \resizebox{0.5\textwidth}{!}{  
    \begin{tabular}{|c|c c |c c c|}
    \hline
    Method & TTO? & Object-Centric? & PSNR $(\uparrow)$  & SSIM $(\uparrow)$  & LPIPS $(\downarrow)$\\
    \hline
    Object-NeRF & Y & Y & 23.01 & 0.841 & 0.118 \\
    MVS-NeRF & N & N & 18.97 & 0.830 & 0.203 \\
    \hline
    COV-NeRF & N & Y & \textbf{25.62} & \textbf{0.905} & \textbf{0.089}\\
    \hline
    \end{tabular}
}
\vspace{-7mm}
\end{center}
\end{table}

\subsection{Sim-to-real Perception}
\label{sec_sim2real_experiments}

\begin{table*}[h!]
\vspace{3mm}
\caption{Sim-to-Real Improvement}
\vspace{-4mm}
\label{table_sim_to_real_robot}
\begin{center}
    \begin{tabular}{c|c|c c|c c}
        &  & \multicolumn{2}{|c|}{Grasp Success Rate $(\uparrow)$} & \multicolumn{2}{c}{Mask AP, modal / amodal $(\uparrow)$} \\
        Method & Real Supervision & Mixed-Clutter & Hard-Specular & Mixed-Clutter & Hard-Specular \\
        \hline
        Pure sim-to-real & None & $0.700\pm0.028$ & $0.463\pm0.035$ & 28.5 / 27.3 & 13.8 / 12.4 \\ 
        \hline
        Finetune segm & 100 scenes, segm & $0.742\pm0.035$ & $0.567\pm0.036$ & 59.4 / 57.3 & 66.5 / 65.3 \\
        CyCADA \cite{cycada} & 100 scenes, rgb & $0.725\pm0.035$ & $0.453\pm0.037$  & 31.2 / 29.2 & 28.1 / 28.0 \\
        DDIB \cite{ddib} & 100 scenes, segm &  $0.727\pm0.031$ & $0.598 \pm 0.037$ & 36.1 / 35.1 & 31.7 / 31.3 \\
        \hline
        COV-NeRF (ours) & 100 scenes (5 segm / 95 rgb only) & $\mathbf{0.929\pm0.021}$ & $\mathbf{0.807\pm0.022}$ & \textbf{62.7 / 61.8} & \textbf{72.9 / 72.4} \\
        \hline
    \end{tabular}
\end{center}
\vspace{-4mm}
\end{table*}

In this section, we study COV-NeRF's effectiveness at improving visual sim-to-real transfer.
Using a real-world bin picking system, we construct challenging configurations that exhibit a sim-to-real gap and evaluate COV-NeRF's effectiveness at reducing the gap.
Our system consists of an ABB 1200, a 6-cup suction gripper, and 6 RGB cameras mounted over a bin.
The robot must grasp objects from the bin one at a time, and transport them to an adjacent bin.

We construct a state-of-the-art bin picking system, using the following components as a representative sample of fundamental perception capabilities for robotic applications:
\begin{itemize}
\item 
Instance segmentation: we modify a SOTA method, MaskDINO \cite{MaskDINO}, to predict both modal and amodal masks.
We train it on COB-3D-v2, which has similar scene composition to our real environment.
\item 
Depth estimation: we train a SOTA model for multi-view stereo, MVS-Former \cite{mvsformer}, on COB-3D-v2.

\item
Grasping: we use a fully-convolutional grasp-quality CNN (FC-GQCNN) \cite{fcgqcnn} in the style of DexNet 3.0 \cite{dexnet3}.
It is trained in simulation to place suction cups on flat surfaces and near the object's center of mass.
For each unoccluded object in the scene (as predicted by MaskDINO), we sample grasps based on a crop of the depth map predicted by MVS-Former.

\end{itemize}

\begin{figure}[b!]
  \centering
  \includegraphics[width=0.48\textwidth]{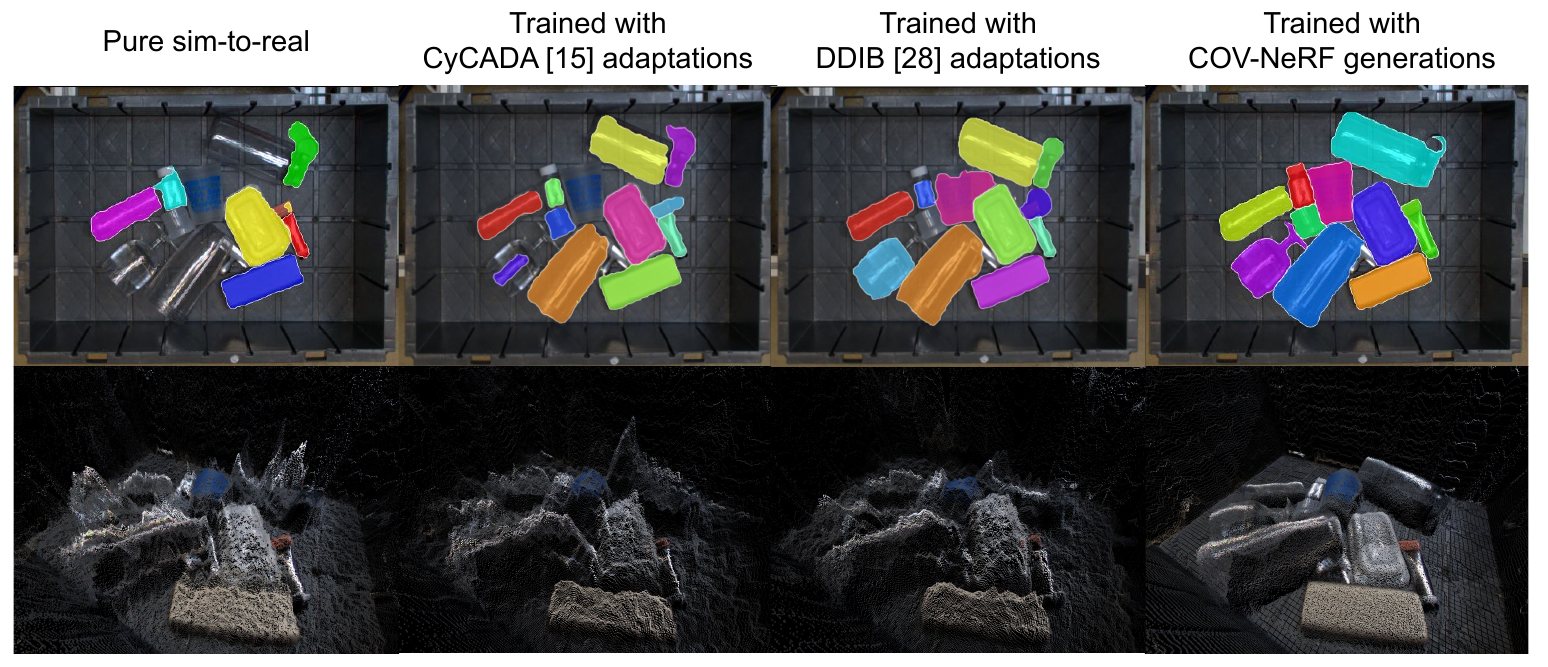}
  \vspace{-7mm}
  \caption{
  Sample instance segmentation predictions from MaskDINO (top row) and stereo depth predictions from MVS-Former (bottom row) resulting from the sim-to-real methods evaluated in Table \ref{table_sim_to_real_robot}.
  COV-NeRF enables substantial improvement in both modalities.
  }
  \label{figure_sim2real_qualitative}
\end{figure}

We consider a few different scenarios:
\begin{enumerate}
\item 
Mixed-Clutter: generic household objects are arranged chaotically inside the bin.
The clutter is generally challenging for both segmentation and depth; it can be especially difficult to reason about occlusions.

\item 
Hard-Specular: we curate a set of challenging transparent and reflective objects.
These non-Lambertian surfaces cannot be accurately sensed by standard depth cameras, necessitating the use of learned methods like MVS-Former.

\end{enumerate}

\begin{figure*}[!t]
  \centering
  \includegraphics[width=0.95\textwidth]{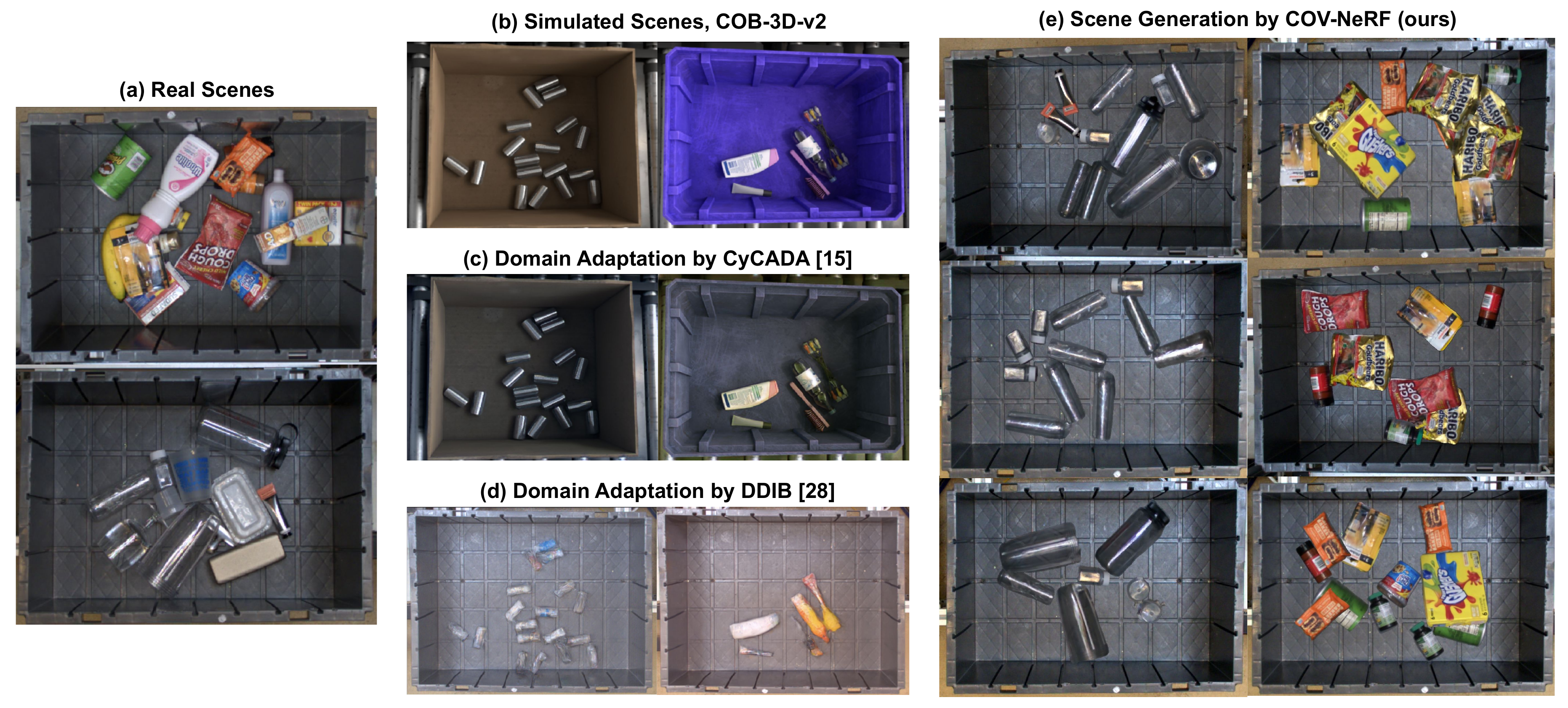}
  \vspace{-2mm}
  \caption{
  (a) Representative real images from the Mixed-Clutter (top) and Hard-Specular (bottom) scenarios.
  (b) Sample simulated scenes from COB-3D-v2.
  (c) CyCADA \cite{cycada} adapts the scenes from (b) to more closely resemble samples from the real world, but its cycle consistency objectives result in only a mild re-styling of the sim scenes.
  (d) DDIB \cite{ddib} produces more visually realistic adaptions of (b), but violates the original scene semantics.
  (e) Instead of adapting simulated scenes, COV-NeRF composes new scenes using explicit object representations extracted from (a).
  }
  \vspace{-4mm}
  \label{figure_synthetic_examples}
\end{figure*}

For each scenario, we first evaluate the baseline system.
As expected, there is substantial room for improvement because it has not seen any real data, let alone these particular objects and environmental conditions.
We then benchmark the following strategies for reducing the sim-to-real gap:
\begin{itemize}

\item 
COV-NeRF: 
We generate a synthetic dataset (following Section \ref{sec_scene_generation}) and use it to finetune MaskDINO and MVS-Former.
Although COV-NeRF generalizes well for view synthesis (as evidenced by Figure \ref{figure_view_synthesis_real}), we found that even tiny amounts of mask supervision vastly improve the synthetic dataset, especially the sharpness of the rendered masks.
We finetune COV-NeRF using 100 real scenes encountered by the baselines system, of which 5 have mask supervision, and use the extracted objects to generate 1000 synthetic scenes.

\item 
CyCADA \cite{cycada}: This well-established domain adaptation method uses GANs to translate images between sim and real, with cycle consistency in both RGB and semantic segmentation.
Deriving semantic segmentation from instance masks using
three classes, $\{\textit{background, object, edge}\}$, we train CyCADA using COB-3D-v2 and 100 real scenes from each scenario.
We then use the trained GANs to augment the training of MaskDINO and MVS-Former: for each simulated scene in COB-3D-v2, we use the original labels as supervision for the adapted images produced by CyCADA.

\item 
Dual Diffusion Implicit Bridges \cite{ddib}: DDIB performs image-to-image translation using diffusion models, which are the modern method of choice for image generation.
Inspired by conditional diffusion models (such as StableDiffusion's img2img and depth2img modes \cite{stablediffusion}), we extend DDIB to be mask-conditioned, and train it to translate between COB-3D-v2 and our real scenarios.
We then use the same procedure as for CyCADA to train MaskDINO and MVS-Former.

\item 
Segmentation Finetuning:  As a baseline, we finetune MaskDINO on real scenes using mask supervision.
Note that this does not affect MVS-Former.
\end{itemize}

In Figure \ref{figure_synthetic_examples}, we visualize the outputs of each method, and in Table \ref{table_sim_to_real_robot}, we quantitatively evaluate grasp success and instance segmentation (mask AP).
Depth estimation can only be evaluated indirectly through grasp success, since we cannot obtain ground-truth in the real world (especially for Hard-Specular), but we study qualitative examples in Figure \ref{figure_sim2real_qualitative}.
With as little as 5 scenes of mask supervision, COV-NeRF enables substantial improvement in both grasp success and mask AP.
CyCADA has almost no effect on the real system: the cycle consistency objective and the challenges of GAN optimization prevent it from substantially altering the simulated images.
DDIB's adaptations have more realistic low-level textural details, but are still vastly less effective than the scenes synthesized by COV-NeRF.

\section{Conclusion}

We presented COV-NeRF, an object-composable and scene-generalizable neural renderer that can the close sim-to-real gap for a variety of perception modalities, including ones where direct supervision cannot be obtained in the real world.
After validating COV-NeRF against existing NeRF methods, we explored its effectiveness at combating sim-to-real mismatch in challenging bin-picking scenarios.
We showed that COV-NeRF can generate targeted synthetic data that is effective at improving state-of-the-art perception methods, translating to significant improvements in end-to-end application performance.

Despite its demonstrated results, COV-NeRF's rendering model suffers from a few limitations.
It does not account for higher-order visual effects like reflections or variations in scene lighting, and it must imagine the appearance of the occluded portions of objects.
We hope to address this in future work, as it would further enhance the diversity and realism of the scenes that COV-NeRF can generate.


\newpage
\begin{thebibliography}{99}
\bibitem{nerf}
Mildenhall, Ben, et al. 
"NeRF: Representing scenes as neural radiance fields for view synthesis." 
European Conference on Computer Vision (ECCV), 2020.

\bibitem{retinagan}
Ho, Daniel, et al. 
"RetinaGAN: An object-aware approach to sim-to-real transfer." 
IEEE International Conference on Robotics and Automation (ICRA), 2021.

\bibitem{pixelnerf}
Yu, Alex, et al. 
"PixelNeRF: Neural radiance fields from one or few images." 
IEEE/CVF Conference on Computer Vision and Pattern Recognition (CVPR), 2021.

\bibitem{mvsnerf}
Chen, Anpei, et al. 
"MvsNeRF: Fast generalizable radiance field reconstruction from multi-view stereo." 
IEEE International Conference on Computer Vision (ICCV), 2021.

\bibitem{nerformer}
Reizenstein, Jeremy, et al. 
"Common objects in 3d: Large-scale learning and evaluation of real-life 3d category reconstruction."
IEEE International Conference on Computer Vision (ICCV), 2021.

\bibitem{osf}
Yu, Hong-Xing, et al. 
"Learning object-centric neural scattering functions for free-viewpoint relighting and scene composition." 
Transactions on Machine Learning Research (TMLR), 2023.

\bibitem{objectnerf}
Yang, Bangbang, et al. 
"Learning object-compositional neural radiance field for editable scene rendering." 
IEEE International Conference on Computer Vision (ICCV), 2021.

\bibitem{panoptic_nf}
Kundu, Abhijit, et al. 
"Panoptic neural fields: A semantic object-aware neural scene representation." 
IEEE/CVF Conference on Computer Vision and Pattern Recognition (CVPR), 2022.

\bibitem{panoptic_lifting}
Siddiqui, Yawar, et al. 
"Panoptic lifting for 3d scene understanding with neural fields."
IEEE/CVF Conference on Computer Vision and Pattern Recognition (CVPR), 2023.

\bibitem{dexnerf}
Ichnowski, Jeffrey, et al. 
"Dex-NeRF: Using a neural radiance field to grasp transparent objects."
Conference on Robot Learning (CoRL), 2021.

\bibitem{evonerf}
Kerr, Justin, et al. 
"Evo-NeRF: Evolving nerf for sequential robot grasping of transparent objects." 
Conference on Robot Learning (CoRL), 2022.

\bibitem{nerf_stereo}
Tosi, Fabio, et al. 
"NeRF-Supervised Deep Stereo."
IEEE/CVF Conference on Computer Vision and Pattern Recognition (CVPR), 2023.

\bibitem{grasp_nerf}
Dai, Qiyu, et al. 
"GraspNeRF: multiview-based 6-DoF grasp detection for transparent and specular objects using generalizable NeRF." 
IEEE International Conference on Robotics and Automation (ICRA), 2023.

\bibitem{domain_randomization}
Tobin, Josh, et al. 
"Domain randomization for transferring deep neural networks from simulation to the real world." IEEE International Conference on Intelligent Robots and Systems (IROS), 2017.

\bibitem{cycada}
Hoffman, Judy, et al. 
"CyCADA: Cycle-consistent Adversarial Domain Adaptation." 
International Conference on Machine Learning (ICML), 2018.

\bibitem{rlcyclegan}
Rao, Kanishka, et al. 
"RL-CycleGAN: Reinforcement learning aware simulation-to-real." 
IEEE/CVF Conference on Computer Vision and Pattern Recognition (CVPR), 2020.

\bibitem{cacti}
Mandi, Zhao, et al. 
"CACTI: A framework for scalable multi-task multi-scene visual imitation learning." 
arXiv preprint arXiv:2212.05711, 2022.

\bibitem{rosie}
Yu, Tianhe, et al. 
"Scaling robot learning with semantically imagined experience." 
arXiv preprint arXiv:2302.11550, 2023.

\bibitem{genaug}
Chen, Zoey, et al. 
"GenAUG: Retargeting behaviors to unseen situations via generative augmentation." 
Robotics: Science and Systems (RSS), 2023

\bibitem{occupancy}
Mishra, Nikhil, et al. 
"Convolutional Occupancy Models for Dense Packing of Complex, Novel Objects."
IEEE/RSJ International Conference on Intelligent Robots and Systems (IROS), 2023.

\bibitem{dexnet3}
Mahler, Jeffrey, et al. 
"Dex-net 3.0: Computing robust vacuum suction grasp targets in point clouds using a new analytic model and deep learning." 
IEEE International Conference on Robotics and Automation (ICRA), 2018.

\bibitem{fcgqcnn}
Satish, Vishal, Jeffrey Mahler, and Ken Goldberg. 
"On-policy dataset synthesis for learning robot grasping policies using fully convolutional deep networks." 
IEEE Robotics and Automation Letters 4.2 (2019): 1357-1364.

\bibitem{MaskDINO}
Li, Feng, et al. 
"Mask dino: Towards a unified transformer-based framework for object detection and segmentation." 
IEEE/CVF Conference on Computer Vision and Pattern Recognition (CVPR), 2023.

\bibitem{mvsformer}
Cao, Chenjie, Xinlin Ren, and Yanwei Fu. 
"MVSFormer: Multi-View Stereo by Learning Robust Image Features and Temperature-based Depth." Transactions on Machine Learning Research (TMLR), 2022.

\bibitem{meshrcnn}
Gkioxari, Georgia, Jitendra Malik, and Justin Johnson. 
"Mesh R-CNN." 
IEEE International Conference on Computer Vision (ICCV), 2019.

\bibitem{neuray}
Liu, Yuan, et al. 
"Neural rays for occlusion-aware image-based rendering." 
IEEE/CVF Conference on Computer Vision and Pattern Recognition (CVPR), 2022.

\bibitem{mujoco}
Todorov, Emanuel and Erez, Tom and Tassa, Yuval.
"MuJoCo: A physics engine for model-based control."
IEEE/RSJ International Conference on Intelligent Robots and Systems (IROS), 2012.

\bibitem{ddib}
Su, Xuan and Song, Jiaming and Meng, Chenlin and Ermon, Stefano.
"Dual Diffusion Implicit Bridges for Image-to-Image Translation."
International Conference on Learning Representations (ICLR), 2023.

\bibitem{stablediffusion}
Rombach, Robin, et al. 
"High-resolution image synthesis with latent diffusion models." IEEE/CVF Conference on Computer Vision and Pattern Recognition (CVPR), 2022.

\bibitem{ddpm}
Ho, Jonathan, Ajay Jain, and Pieter Abbeel. 
"Denoising diffusion probabilistic models." 
Advances in Neural Information Processing Systems (NeurIPS), 2020.

\bibitem{cutpaste}
Ghiasi, Golnaz, et al. 
"Simple copy-paste is a strong data augmentation method for instance segmentation." 
IEEE/CVF Conference on Computer Vision and Pattern Recognition (CVPR), 2021.

\bibitem{data_dreaming}
Sieb, Maximilian, and Katerina Fragkiadaki. 
"Data dreaming for object detection: Learning object-centric state representations for visual imitation." 
IEEE-RAS International Conference on Humanoid Robots (Humanoids), 2018.


\end{thebibliography}
\end{document}